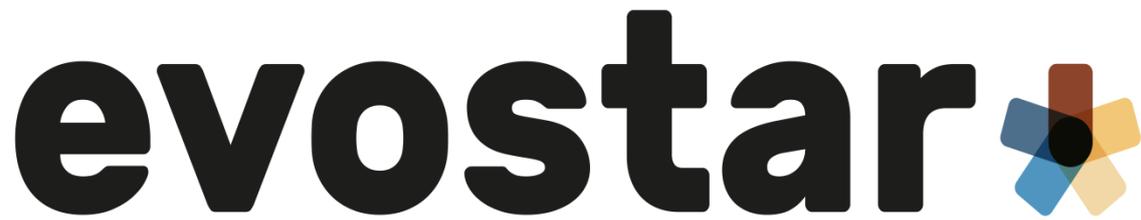

# Evo* 2020

## The Leading European Event on Bio-Inspired Computation

*Online Conference. 15-17 April 2020*

http://www.evostar.org/2020/

# – LATE-BREAKING ABSTRACTS –

**Editors:**

A.M. Mora
A.I. Esparcia-Alcázar

# Preface

This volume contains the Late-Breaking Abstracts submitted to the Evo* 2020 Conference, that took place online, from 15 to 17 of April.

These papers where presented as short talks and also at the poster session of the conference together with other regular submissions.

All of them present ongoing research and preliminary results investigating on the application of different approaches of Evolutionary Computation and other Nature-Inspired techniques to different problems, most of them real world ones.

We consider these contributions as very promising, since they outline some of the incoming advances and applications in the area of nature-inspired methods, mainly Evolutionary Algorithms.

*Antonio M. Mora*
*Anna I. Esparcia-Alcázar*

# Table of contents

# A Multi-strategy LSHADE Algorithm and its Applications on Temporal Alignment

Zhenglei Wei[1][0000-0002-2530-8365] , Changqiang Huang

[1] Air Force Engineering University, Xi'an, Shannxi, China
zhenglei_wei@126.com

## 1 Introduction

In general, the approaches which can make DTW [1] faster include abstracting the data, indexing for near neighbors search application, pruning the computation of DTW and reducing the alignment path search space. In this paper, a novel alignment method which is combined with a novel hybrid optimization algorithm is proposed to reduce the calculation cost of alignment. Our contributions can be divided into below several parts:

(i) A multi-strategy LSHADE (MLSHADE) algorithm which combines improved LSHADE with adaptive CMA-ES is presented.

(ii) Combined with multi-strategy LSHADE, DTW based on MLSHADE, called MLDTW, is proposed to reduce the alignment complexity and improve the accuracy of alignment.

(iii) The performance of MLDTW is tested in sinusoidal signal and UCR time series archive, which is compared with other known alignment algorithms.

## 2 MLSHADE

Based on the ELSHADE-SPACMA, a novel algorithm, called multi-strategy LSHADE, is proposed. The main framework is divided into two phases, which include LSHADE-SPACMA [2] phase and AGDE [3] phase. The main framework is described in [3]. In order to improve the performance of MLSHADE, three strategies are proposed as follows:

(i) Weighted mutation strategy. To enhance the diversity of population, we propose a weighted mutation strategy, called *current-to-pbest-w/1*.

(ii) Inferior solution search strategy. According to fitness values of solutions, the performance rank $P_{rank}$ can distinguish the superior or inferior solutions. For the $i^{th}$ solution, if $P_{rank}(i) < 0.5$, it is regarded as the superior solution and use the original technique of CMA-ES to generate a new vector. If $P_{rank}(i) > 0.5$, the individual represents the inferior solution, which can be employed to enhance the exploration ability of CMA-ES method. The inferior solutions can be applied to two different update states. More details are presented as follows: *a)*. In state 1, new candidate is generated by superior and inferior solutions in eigen coordinate. *b)*. State 2 updates the shifted mean value by utilizing the difference between the inferior and superior solutions.

(iii) Eigen Gaussian random walk strategy. In the second phase, a Gaussian random walk and the eigen coordinate system are presented to improve the exploitation performance of the AGDE.

$$x_i^{eig,G+1} = Gaussian\left(x_{pbest}^{eig,G}, \sigma\right) + rand_1 \cdot x_{pbest}^{eig,G} - rand_2 \cdot x_r^{eig,G}, \quad rand_1, rand_2 \sim U(0,1) \tag{15}$$

$$\sigma = \left|x_{pbest}^{eig,G} - x_{pworst}^{eig,G}\right| \tag{16}$$

where $x_r^{eig,G}$ is selected randomly from the middle *NP*-2*(100*p*%) at generation *G* under the eigen coordinate. $x_{pbest}^{eig,G}$ and $x_{pworst}^{eig,G}$ are chosen randomly from the best and worst 100*p*% under eigen coordinate.

## 3   Temporal Alignment based on MLSHADE

Combined with multi-strategy LSHADE, the DTW based on MLSHADE, called MLDTW, is proposed to reduce the complexity of alignment and improve the accuracy of alignment.

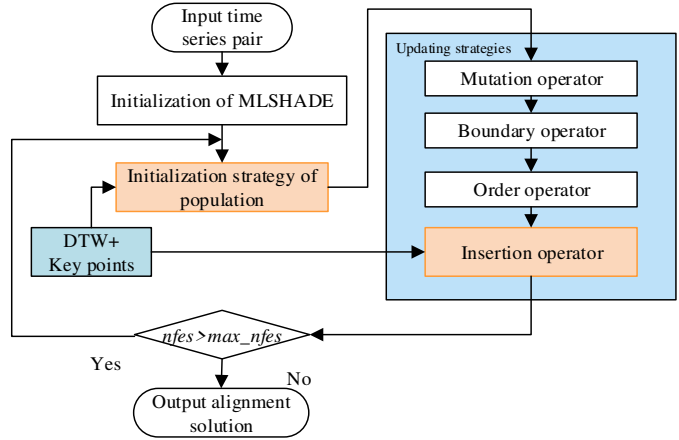

**Fig. 1.** Flow chart of MLSHADE

(i) As a path optimization problem, the variable length encoding is used by MLDTW.

(ii) In MLDTW, the energy function can be thought as fitness function to minimize the following formula.

$$J(p_x, p_y) = \left\|X_p W_x - Y_p W_y\right\|_F^2 \tag{19}$$

where $[p_x^T, p_y^T]^T$ is marked as alignment solution.

(iii) For the novel initialization technique, the key points will be selected to segment the whole alignment path. According to relationship between segmentation points, DTW with short time series segments will be computed.

(iv) For individuals with variable length encoding, we propose a variant of mutation operator to overcome the difference. Meanwhile, boundary operator and order operator are proposed to ensure that the solution satisfies boundary condition and monotonicity condition of DTW. Moreover, to meet to continuity condition, the insertion operator is introduced.

Through combining above strategies, MLSHADE can be illustrated in Fig.1.

## 4 Experiments

This section presents the experiments on temporal alignment based on MLDTW proposed in this paper. For alignment experiments, sinusoidal signal and UCR dataset are utilized.

(i) Sinusoidal signal alignment.

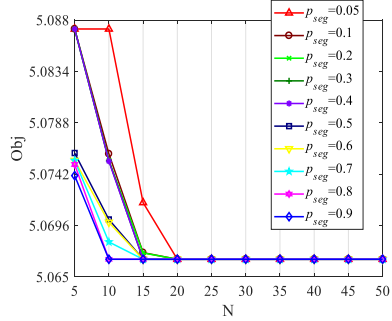

(a) Objective function value across different $N$ and $P_{seg}$ levels.

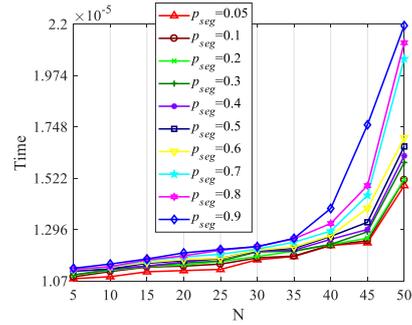

(b) Time results across different $N$ and $P_{seg}$ levels.

**Fig.2**. Alignment results of proposed method across different $N$ and $P_{seg}$ levels.

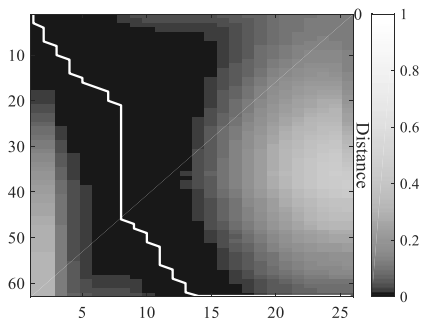

(a) Optimal alignment path based on MLDTW.

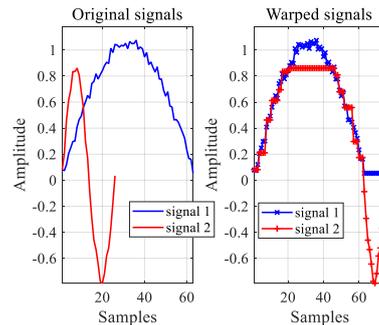

(b) Comparison between original signals and warped signals.

**Fig.3**. Optimal alignment paths and warped signals.

To test and verify the availability and robustness of MLDTW, synthetic sinusoidal signal is employed to construct the alignment problem. The two temporal sinusoidal signals generated with zero-mean Gaussian noise can be shown in left of Fig.3(b).

To test the robustness of MLDTW and the sensitivity of parameters, alignment experiments based on MLDTW are carried out on different population size $N$ and segment rate $P_{seg}$ levels. Alignment results of MLDTW across different $N$ and $P_{seg}$ levels are shown in Fig.2. According to the robustness analysis, the appropriate parameters can be selected, including to $N$=20 and $P_{seg}$=0.1. Based on suitable parameters, optimal alignment path based on MLDTW is shown in Fig.3(a). According to alignment solution, the warped signals are presented in Fig.3(b).

In order to prove the superiority of MLDTW, the comparison experiments are carried out. The experiment results are shown in Table 5, including objective value mean, SD and average time. As can be seen in Table 5, MLDTW which ranks the first in mean possesses the similar performance as ELDTW and NRO.

**Table 1**. Experimental results evaluated by compared alignment methods.

|  | DTW | ELDTW | iLSHDTW | jSODTW | GEDGWODTW | NRODTW | MLDTW |
|---|---|---|---|---|---|---|---|
| Mean | 4.9995 | **4.8988** | 5.8309 | 6.1630 | 5.5078 | **4.8988** | **4.8988** |
| SD | 0.3952 | 0.4246 | 0.8793 | 3.8249 | 0.6151 | **0.3735** | **0.3735** |
| Mean rank | 4 | 3 | 6 | 7 | 5 | 1 | 1 |
| Average Time/s | 2.45e-04 | 1.84e-05 | 1.46e-05 | **1.32e-05** | 3.49e-05 | 2.33e-05 | 1.33e-05 |
| Time rank | 7 | 4 | 3 | 1 | 6 | 5 | 2 |

(ii) Temporal alignment on UCR dataset

To further highlight the effectiveness of proposed alignment measure, the 6 datasets from UCR time series database using the one-nearest-neighbor (1-NN) error rate are employed. Based on parameters obtained from above section, MLDTW is compared against the DTW, ELDTW, iLSHDTW, jSODTW, GEDGWODTW and NRODTW using 1-NN error rate in Fig.8. It is clear that MLDTW outperforms the other alignment methods.

Additionally, the statistical results evaluated by compared alignment methods on UCR database are given in Table 6. The statistical results show that Error Rate (ER) and mean time of MLDTW rank the first and second, respectively.

## 5 Conclusions

As a novel alignment technique, MLDTW using multi-strategy LSHADE and DTW is described. MLSHADE is a novel optimization algorithm, which employs weighted mutation strategy, inferior solution search strategy and Eigen Gaussian random walk strategy at the framework of ELSHADE-SPACMA. In order to improve the efficiency of DTW, a novel alignment approach is proposed by using MLSHADE operators. The analysis of MLDTW is based on sinusoidal signal and six datasets of UCR time series. The statistical results show that MLDTW exerts the characteristics of high accuracy and efficiency compared with other alignment techniques.

Future work can address two points. First, MLDTW should be applied to specific real-world applications. Second, MLDTW can be extended to multi-dimensional time series alignment problems.

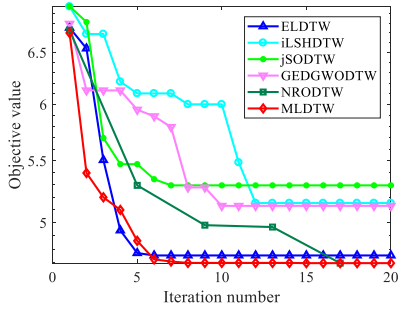

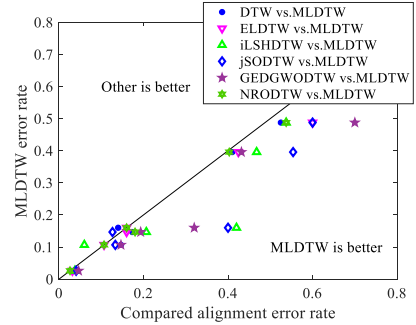

**Fig.7**. Fitness curves of compared methods.  **Fig.8**. 1NN error rates of proposed different alignment measures on UCR database.

**Table 2**. Experimental results evaluated by compared alignment methods on UCR database.

| Type | | DTW | ELDTW | iLSHDTW | jSODTW | GEDGWODTW | NRODTW | MLDTW |
|---|---|---|---|---|---|---|---|---|
| BME | ER | 0.1467 | 0.1067 | **0.0600** | 0.1333 | 0.1467 | 0.1067 | 0.1067 |
| | Time | 2.82e-04 | 7.13e-05 | 1.45e-04 | **5.17e-05** | 7.78e-05 | 6.40e-05 | 6.01e-05 |
| Chinatown | ER | 0.0435 | 0.0319 | 0.0377 | 0.0406 | 0.0467 | **0.0261** | **0.0261** |
| | Time | 4.19e-05 | 8.05e-06 | 1.40e-05 | **5.95e-06** | 7.31e-06 | 6.87e-06 | 6.39e-06 |
| DistalPhalanxTW | ER | 0.4101 | 0.4245 | 0.4676 | 0.5540 | 0.4317 | 0.4029 | **0.3957** |
| | Time | 7.56e-05 | 2.97e-05 | 5.96e-05 | **2.44e-05** | 3.28e-05 | 2.58e-05 | 2.32e-05 |
| DodgerLoopDay | ER | 0.5250 | 0.6000 | 0.5375 | 0.6000 | 0.7000 | 0.5375 | **0.4875** |
| | Time | 1.43e-03 | 3.57e-04 | 7.56e-04 | **2.42e-04** | 3.36e-04 | 3.28e-04 | 3.10e-04 |
| ShakeGestureWiimoteZ | ER | **0.1400** | 0.1600 | 0.4200 | 0.4000 | 0.3200 | 0.1600 | 0.1600 |
| | Time | 4.95e-04 | 1.62e-04 | 3.44e-04 | **1.07e-04** | 1.54e-04 | 1.46e-04 | 1.40e-04 |
| SmoothSubspace | ER | 0.1733 | 0.1600 | 0.2067 | **0.1267** | 0.1933 | 0.1800 | 0.1467 |
| | Time | 3.88e-05 | 5.67e-06 | 8.36e-06 | **3.64e-06** | 4.22e-06 | 3.88e-06 | 4.73e-06 |
| Ranks | ER | 4 | 3 | 5 | 6 | 7 | 2 | **1** |
| | Time | 7 | 5 | 6 | **1** | 4 | 3 | 2 |

## References

1. Bagnall, A., Lines, J., Bostrom, A., Large, J., Keogh, E.: The great time series classification bake off: a review and experimental evaluation of recent algorithmic advances. Data Min. Knowl. Discov. (2017). https://doi.org/10.1007/s10618-016-0483-9.
2. Mohamed, A.W., Hadi, A.A., Fattouh, A.M., Jambi, K.M.: LSHADE with semi-parameter adaptation hybrid with CMA-ES for solving CEC 2017 benchmark problems. In: 2017 IEEE Congress on Evolutionary Computation, CEC 2017 - Proceedings (2017). https://doi.org/10.1109/CEC.2017.7969307.
3. Hadi, A. A, Wagdy A., Jambi K.. Single-objective Real-parameter Optimization: Enhanced LSHADE-SPACMA Algorithm. Tech. Rep., 2018. doi: 10.13140/RG.2.2.33283.20005.

# From Business Curated Products to Algorithmically Generated[*]

Vera Kalinichenko[1] and Garima Garg[2]

[1] University Of California Los Angeles, USA
[2] `vira@g.ucla.edu, kalin.vera@gmail.com`
https://www.uclaextension.edu/digital-technology/data-analytics-management
[3] FabFitFun Inc, West Hollywood, California, USA
`garima.garg@fabfitfun.com`

**Abstract.** This document outlines how at FabFitFun we have developed a bundle of machine learning models that enabled us algorithmically assemble future boxes that are shipped to our members. FabFitFun is a technology and e-commence startup that works with small vendors to discover cool products in a similar fashion to Amazon marketplace. The team of data scientists utilize our historical data which includes details of our customers and products we have sold and are planning to sell, programmatically discover new products. We use classical machine learning supervised models to classify our products and latest technologies like genetic programming, linear optimization to help business to curate products to members in a new era. Genetic programming allows us to spot repetitive features for our supervised learning models, provide product diversity via measuring how far away a newly generated collection of products are from the historically seen and the optimal solution. Our paper demonstrates novel way to evaluate algorithmically created boxes using Sharpe Ratio Concept from finance. Thus, we discover novel products based on the distance threshold from the optimal solutions and adapted way to evaluate boxes with Sharpe Scores. Utilize genetic programming to generate synthetic data, provide diversity, novelty across our products.

**Keywords:** Genetic programming · optimal solutions · novelty · classifier

## 1 Introduction. Box Definition at FabFitFun. Data we collect and use.

Majority of products never repeat, thus that we are faced with a cold start problem. We are addressing this problem via genetic programming and feature enrichment from text. What we do - we provide lifestyle membership for women, we help to exchange ideas within community of women that love fashion, beauty and home decor. Throughout the whole article we speak about assembling a

---

[*] Supported by organization FabFitFun Inc.

EVO* 2020 - Proceedings in ArXiv - Late-Breaking Abstracts

box. Let's define a box as a collection of products. We enable and encourage brand discovery for smaller businesses if they partner with us. Every season our members receive a box with unique set of product items from various categories of beauty, fashion, fitness, home, lifestyle and more. Our products usually do not repeat from one shipped box to another, this provides a challenge to use supervised learning techniques. Since we need to capture product characteristics that are repetitive enough to help distinguish a winner product and consequently a winner box from neutral or so so products. We utilize Elastic Search Engine to parse text into token and convert text tokens to numerical data. Then we use this numerical data as feature enrichment to improve our classifiers.

### 1.1   Flow Diagram of our Process

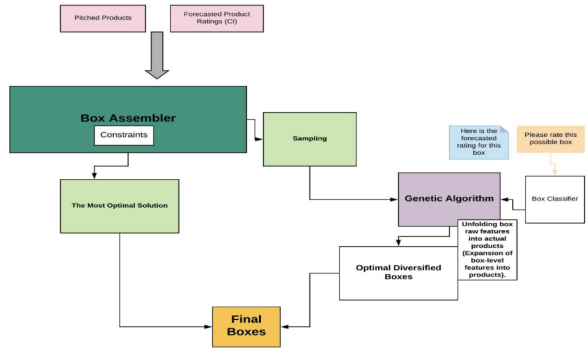

**Fig. 1.** Box Assembly Process

We have gathered and analyzed our historical data that includes customer satisfaction survey feedback, product data and box feedback data provioded back by our clients.

### 1.2   Box Classifier Probability as a Fitness Function for Genetic Algorithm.

We build a data set that contains historical purchase product data by FFF members. At FabFitFun we collect not just our member digital activity but we run several surveys in which we request our customers to rate and provide written reviews of the products our members had purchased and the corresponding boxes.We implemented and trained a classical XGBoost Classifier that is able to distinguish a good box from the bad box (bad box is scored below or equal to 2, and 3 is a neutral box). We use this classifier as base foundation in our genetic programming approach. When we produce programmatically generated boxes (feature collections that uniquely corresponds to physical products) to send the generated box the the classifier to determine its probability to be a good box.

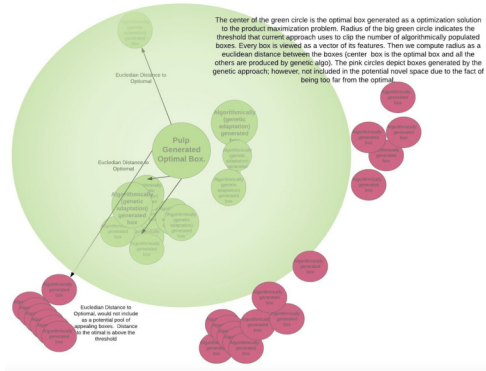

**Fig. 2.**

### 1.3 Solving for the Optimal Box.

A FFF box is a collection of $n$ different products that is selected from a list of $m$ pitched products. We get an optimal box by satisfying these constraints below:

1. Total Retail Value of Box is above $250
2. Total Box cost for FabFitFun is below $ $COGS$ (where $COGS$ is a constant, its value is set by business, not disclosed here)
3. Diversity in product categories in the Box, we want to have at least one product from each of the categories $C_1$, $C_2$ and $C_3$

Our objective function is to maximize ratings of the future shipped boxes. We able to solve for the optimal box based on the predictive ratings computed by one of our multi-class classifiers. Every future product rating is a predicted product rating based on historical products and their features, and label historical data. Every customer rates a product, we collect that data. Every product is a collection of features such as price, cost of production, size of the item, category of the item, sub-category of the item, etc..We are able to train on the data and successfully predict the future rating.

### 1.4 Genetic Programming. Ability to Generate Synthetic Data.

Our Algorithm:

1. View every historical box as a collection of features (genes). Data clean up and quality checks.
2. Data Generation Process
3. Specify size of population (number of boxes to keep in every generation), and number of generations to go on generating new data.
4. Ability to pick 2 or more parents from the prior generation.
5. Ability generate new box by specifying algorithmically which features will be taking from which parent

4    Vera Kalinichenko and Garima Garg6. Implementing fitness function as a probability higher than certain threshold (more than 0.5) of the populated box being a "good" box (high rated box)

To ensure that we can provide enough diversity and novelty in our "next generation" boxes. This approach specifically drives from the historical data and measures how novel is the new box from the historical boxes in terms of list of categories and other features. We are specifically focus not only on the optimal solution that optimize towards high score on the box but have an ability to move away from the optimal solution in order to ensure novelty and discovery of product features that we have not seen in the past.

## 2  Box Evaluation

We need to design and implement the metric which will quantitatively describe the goodness of the generated box. Our proposal borrowed from finance portfolio evaluation idea, Sharpe measure, will be adapted here at FabFitFun as follows. Consider k-th generated box with its corresponding product ratings $b_k = (r_{k1}, r_{k2}, r_{k3}, r_{k4}, r_{k5}, r_{k6}, r_{k7}, r_{k8})$ where k - is a positive integer and ranges from 1 to N (fixed number of programmatically generated boxes). We define by Rf our benchmark, the best box rating from the prior season. This is our ground truth. In this case the evaluation metric always will rely on the most recent best box rating available and to ensure constant lift in generated data.

$$(1) S_k = \frac{R_k - R_f}{\sigma_{prior}} \tag{1}$$

where $R_k$ is the rating of the $k-$th algorithmically assembled box.

$$(2) R_f = \max_{i=1}^{M} R_i \tag{2}$$

where $M$ is number of unique boxes shipped to FabFitFun members in the prior season, and $R_i$ is the rating of the $i-$th shipped box where

$$\sigma_{R_k} = \sqrt{\left(\frac{1}{N_k - 1} \sum_{1}^{N_k} (r_i^k - \overline{r^k})\right)} \tag{3}$$

and $N$ number of programmatically generated boxes via GA and Box Assembly.

# Plant Propagation & Hard Hamiltonian Graphs

Joeri Sleegers[1][0000−0003−1701−6319]
Daan van den Berg[1][0000−0001−5060−3342]

Universiteit van Amsterdam
joeri.sleegers@student.uva.nl, d.vandenberg@@uva.nl

**Abstract.** Although the Hamiltonian cycle problem is known to be NP-complete, only a few graphs are actually hard to decide for complete backtracking algorithms running on large ensembles of random graphs. Historically, these hard instances are found near the Komlós-Szemerédi bound, the average vertex degree where the Hamiltonian probability phase transition occurs. In this preliminary investigation, we take a different approach, generating hard graphs with two evolutionary algorithms. We find completely new and counterintuitive results.

**Keywords:** Hamiltonian cycle problem · instance hardness · phase transition · evolutionary algorithms · plant propagation algorithm

## 1 The Hamiltonian cycle problem

The undirected Hamiltonian cycle problem involves deciding whether a given graph of $v$ vertices and average degree $d$ contains a closed path that visits every vertex exactly once. Known to be NP-complete, quite a few complete algorithms exist for the problem, but none of those runs in subexponential time. The dynamic programming Help-Karp algorithm is quite memory intensive, but by $O(n^2 \cdot 2^n)$ still holds the lowest time complexity [3]. Depth-first based algorithms such as Cheeseman's, Van Horn's, Rubin's, and Vandegriend-Culberson's are far more memory efficient, but take more time in the theoretical worst case: $O(v!)$ [1][4][7][12]. Even for the least sophisticated of these algorithms however, the majority of randomly generated graphs is relatively easily decided. Low-degree graphs require few recursions, so an exhaustive search is quickly completed. High-degree graphs however, contain many Hamiltonian cycles, so one is easily found. The hardest graphs reside in between, right around the Komlós-Szemerédi bound of average degree $v \cdot ln(v) + v \cdot ln(ln(v))$ edges, where the probability of a random graph being Hamiltonian goes from almost zero to almost one as $d$ increases [5].

More sophisticated backtracking algorithms such as Vandegriend-Culberson's comb out many of these hard graphs using early-decision techniques, clever pruning and sensible next-vertex heuristics. For this study, we will use a backtracking algorithm that prioritizes low-degree vertices over high-degree vertices, deploys two edge pruning techniques (path pruning and neighbor pruning) and several checks for non-Hamiltonicity (such as degree-one nodes). This algorithm, which inherits most of its techniques directly from Vandegriend-Culberson's, outperforms all aforementioned backtracking algorithms on large ensembles of randomly generated graphs (results are to be published). For precise algorithmic details, we'll refer the reader to our open-source repository [11].

But even though all these advances reduce the average decision time dramatically, the hardest graphs still reside around the Komlós-Szemerédi bound in these random ensembles. In this preliminary investigation, we take a different approach, 'optimizing' graphs to be as hard as possible for this algorithm. Remarkably enough, the resulting graphs, which are indeed very hard, reside in a totally different part of the combinatorial state space.

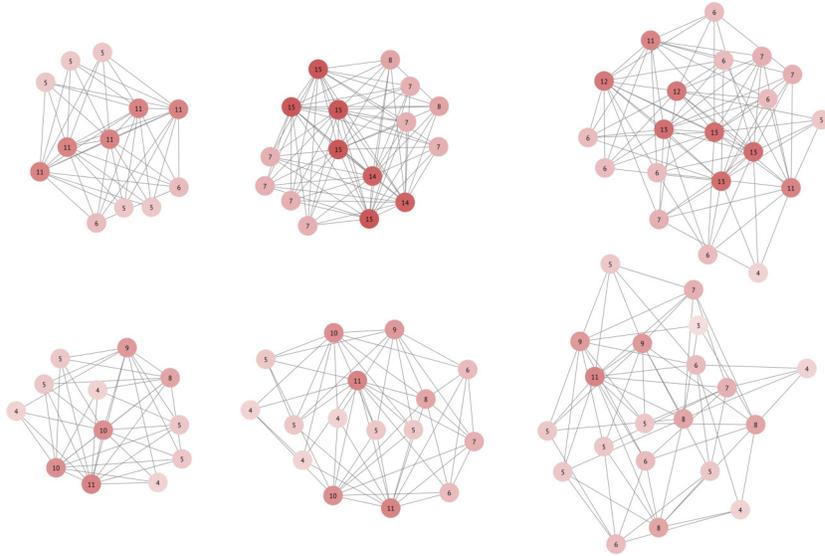

**Fig. 1.** The hardest Hamiltonian graphs of 12, 16 and 20 vertices found by a hillclimber (top row) and a plant propagation algorithm (bottom row). Graphs appear to converge to a 'hamburger structure', with a 'fat layer' of high degree vertices in the middle, flanked by two 'lightweight' layers of low degree vertices on either side.

## 2   Two Evolutionary Algorithms

We use two evolutionary algorithms to make hard graphs: a hillclimber and a plant propagation algorithm [10][8]. For both, the fitness (or objective value) is given by the number of recursions needed by the backtracking algorithm described in Section 1. So the more recursions, the longer the decision time, the harder the graph, and the higher its fitness.

The hillclimber starts off from a random graph of $v = \{12, 16, 20\}$ vertices, and edge degree $d = \lfloor v \cdot ln(v) + v \cdot ln(ln(v)) \rfloor$. Its inception thereby lies exactly on the Komlós-Szemerédi bound, where we would historically expect the hardest graphs to reside [1][4][12]. Each iteration, one of three possible mutation types is randomly chosen and applied: to **insert** an edge at a random unoccupied place in the graph, to randomly **delete** an existing edge from the graph, and to **move** and edge, which is effectively equal to a delete-mutation followed by an insert-mutation (on a *different* unoccupied place). The mutated graph is kept iff fitter, and the mutation is reverted otherwise. An important observation herefrom is that the graphs do not necessarily retain their initial connectivity throughout the evolutionary process, which is somewhat uncommon in this kind of study. But it's exactly this relaxation that provides us with some surprising results.

The plant propagation algorithm is a population-based evolutionary algorithm that can be applied to a broad spectrum of continuous, discrete and mixed objective landscapes in scientific, industrial and even artistic optimization problems [8][9][2][13][6]. To meet these different requirements, various adaptations have been implemented but the core of the algorithm is always the same: a population of solutions from which fitter individuals spawn many offspring with few mutations, and unfitter individuals spawn few offspring with many mutations, all in an effort to balance the powers of exploration and exploitation in a problem's state space.

The version in this experiment is most closely related to the variants used for optimizing the traveling salesman and timetabling problems [9][2]. Maintaining a constant-sized population of 10 individuals (i.e. undirected graphs) descendingly sorted to fitness (i.e. number of recursions needed to decide the graph), the two fittest individuals each spawn five offspring (i.e. new graphs) which all receive one random mutation. Iff any of these offspring is fitter than its parent, it replaces it. The eight unfitter individuals each spawn one offspring which receives 20 mutations, and again replaces its parent when fitter.

## 3 Experiment and Results

Both algorithms get 30 random initializations, 10 for each $v \in \{12, 16, 20\}$ and a corresponding $e \in \{15, 23, 31\}$ – exactly on the Komlós-Szemerédi bound – after which they are run for exactly 500 function evaluations of either evolutionary algorithm. These numbers might appear small, but as we are pushing towards the bounds of an NP-complete problem, a single evaluation can easily take up millions of recursions, even for graphs this small.

Both algorithms find hard graphs in all size categories. In most ensembles, more than half of the evolutionary runs generates graphs requiring over 10,000 recursions, sometimes even ranging in the millions (Table 1). The hillclimber appears to outperform plant propagation in finding hard graphs which is in some sense remarkable as it is prone to getting stuck in local maxima. This might indicate that the graph hardness landscape is largely convex, but it is also not unthinkable that in longer runs, PPA would eventually outperform the hillclimber, as has been witnessed before [2]. What is quite remarkable though, is that all configurations appear to converge to a 'hamburger structure', with approximately 40% 'fat' nodes of high degree being sandwiched between two layers of 30% 'light' nodes of low degree. (Fig. 1).

But what is even more remarkable, is that these graphs all have an edge degree that lies considerably higher than the Komlós-Szemerédi bound where previous investigations by Cheeseman et al., Van Horn et al. and Vandegriend & Culberson found the hardest graphs. There might be several not-so-trivial explanations for this.

## 4  Discussion

A first explanation for these surprising results is that these results are specific for the backtracking algorithm we used. This is unlikely however, as the algorithm minimizes most other backtracking algorithms found in literature (yet unpublished results). Put differently: chances are very high that these graphs are *also* hard for other complete backtracking algorithms but evidence pending, this possibility can not be completely ruled out.

A second explanation might be that in most studies on Hamiltonian cycle backtracking algorithms, runs are cutoff after a preset number of recursions, as even small graphs can take up significant decision time. However, these cutoff points are usually situated near the Komlós-Szemerédi bound, and not in edge-dense regions where the 'hamburger graphs' would be located.

The most tempting explanation might therefore come from the fact that on first glance, these graphs have low Kolmogorov complexity – they are *structured*. As unstructured objects in any randomly generated ensemble vastly outnumber the structured objects, the chances of being created by a stochastic process (which is the case in most large-scale comparative studies) are microscopic. One would simply not find them unless knowing where to look. These graphs are an isolated island of structured hardness in an ocean of unstructured ease. Whether more such islands exist, and what they look like, awaits further exploration.

**Table 1.** Both the hillclimber and plant propagation are succesful in finding hard-to-decide graphs within 500 function evaluations on 6x10 graphs of different sizes $v$. The two last columns are the number of graphs which required over 10,000 recursions, their average degree, and in brackets their *expected* average degree based on the Komlós-Szemerédi bound.

| size | algorithm | max recursions | avg recursions | over 10K | avg deg |
|------|-----------|----------------|----------------|----------|---------|
| 12 | HC | 61,051 | 36,413 | 6/10 | 7.61 (2.56) |
| 12 | PPA | 12,479 | 3,395 | 2/10 | 6.67 (2.56) |
| 16 | HC | 198,557,095 | 19,908,439 | 6/10 | 7.84 (2.84) |
| 16 | PPA | 211,475 | 43,475 | 5/10 | 6.38 (2.84) |
| 20 | HC | 99,742,171 | 10,063,317 | 7/10 | 6.47 (3.05) |
| 20 | PPA | 1,130,923 | 357,725 | 8/10 | 6.08 (3.05) |

# Simplified Paintings-from-Polygons is NP-Hard

Daan van den Berg[0000−0001−5060−3342]

University of Amsterdam
d.vandenberg@uva.nl

**Abstract.** The simplified paintings-from-polygons problem (SPFP), in which paintings or other digital images are approximated by heuristically arranging overlapping semi-opaque colored polygons, is NP-hard. Every instance of the subset sum problem can be transformed to a SPFP instance, solved by some algorithm, and transformed back. Whichever algorithm one chooses, it cannot be more efficient than the most efficient algorithm for subset sum. Since subset sum is known to be NP-hard, and SPFP is at least equally hard, SPFP must also be NP-hard.

**Keywords:** Paintings-from-Polygons · PFP · Alpha Compositing · NP-hard · Evolutionary Algorithms · Plant Propagation Algorithm

## 1 Introduction

Being a blog topic of artistic nature for quite some time, approximating paintings from optimally arranging a limited set of semi-opaque colored polygons has been elevated to the realm of science since EvoMusArt 2019 [1] [3]. Having both a strong visual appeal and an untraversably large combinatorial state space, the problem proves an appealing testing ground for heuristic algorithms such as hill climbing, simulated annealing and the plant propagation algorithm [4][6][1][5]. Starting from an initial configuration of randomly scattered semi-opaque (partially) overlapping polygons, the various algorithms applied four mutation types: moving a polygon's vertex, changing a polygon's color and/or opacity, transferring a vertex from one polygon to another, and changing the 'drawing index' – assigning which polygon is to be drawn first, which second, and so forth during the rendering process. Numbers of vertices and polygons were constant throughout the experiments, and the change in pixel-by-pixel difference between the rendered polygon constellation and the target image, quantified in the mean squared error (MSE), monitored the progress for the different optimization algorithms (Fig. 2).

However visually appealing, no formal proof of the problem's hardness was given by the authors. To make some headway in this direction, I will show that a simplified version of this problem is NP-hard. The simplification is rather strict: given a number of 50% opaque greyscale polygons and a target greyscale image, the objective is to approximate any single pixel within the target image as closely as possible. Apart from some practicalities and common sense requirements (the number of vertices per polygon should be $\geq 3$, there should be more than 0 polygons etc.) the problem is as general as possible. A key role in the proof is played by the polygons' opaqueness and their drawing index – the order in which they are rendered to a 'flat' .png image, a process known as *alpha compositing*. For every pixel in a polygon constellation, the lowest level is given by the background color of the canvas, which is always black, solid, and fully opaque:

$$color_0 = 0 \qquad (1)$$

---
[1] Conference track of EvoStar 2019 in Leipzig, Germany

Then, the color of the pixel is sequentially updated for each polygon covering that pixel. This can be recursively formulated as

$$color_i = \alpha_i \cdot color_i + (1 - \alpha_i) \cdot color_{i-1} \qquad (2)$$

in which $i$ indicates the $i^{th}$ polygon covering the pixel, and $\alpha_i$ its opaqueness. Note that therefore, the influence of the lastly drawn polygon usually outweighs the influence of earlier drawn polygons on the rendered pixel's final color. For SPFP, all polygons have fixed 50% opaqueness, simplifying equation 3 to

$$color_i = \frac{1}{2} color_i + \frac{1}{2} \cdot color_{i-1} \qquad (3)$$

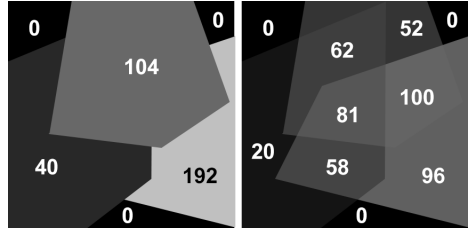

**Fig. 1. Left:** alpha-compositing of three fully opaque ($\alpha = 1$) greyscale polygons on a black canvas. **Right:** the same polygons, drawn in the same order, but in half opaqueness ($\alpha = 0.5$). Numbers inside the areas are greyscale values.

So if a polygon constellation holds seven polygons with greyscale colors (192, 40, 104, 16, 85, 17, 50) of which the first three are chosen to cover a pixel, the pixel's rendered color would be

$$\frac{1}{2} \cdot 104 + \frac{1}{4} \cdot 40 + \frac{1}{8} \cdot 192 = 81 \qquad (4)$$

(Fig. 1 contains this exact numerical example in the central area). After sequentially rendering all the polygons in the constellation, the resulting 'flat' .png

Simplified Paintings-from-Polygons is NP-Hard 3

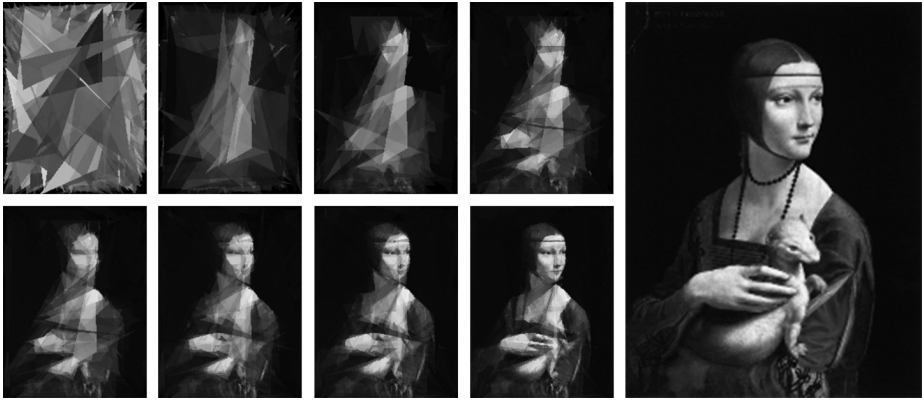

**Fig. 2.** The (simplified) paintings-from-polygons problem ((S)PFP) involves approximating a target image, usually a painting, from heuristically rearranging a set of semi-opaque polygons. The objective is to minimize the pixel-by-pixel error (MSE) between the rendered polygon constellation (smaller subfigures LRTB) and the target image (large subfigure). Numbers of vertices and polygons are fixed throughout the run.

image can then be MSE-wise compared to the target image. Better (heuristic) algorithms obtain ever lower MSE-values throughout the run, arranging polygon constellations ever more towards the target image (Fig. 2).

## 2   NP-Hardness: From Subset Sum to SPFP

In its optimization form, subset sum is the task of approximating a target value $t$ as closely as possible by adding up a number of elements from a set $S$ of $m$ given integers $v_1...v_m$, for example $S = \{13, 17, 21, 23\}$ with $t = 41$. The problem is known to be NP-hard, which means there is no known exact algorithm that runs in subexponential time [2]. In its decision form ("which of the integers $v$ sum up *exactly* to $t$ ?"), it is NP-complete because of its polynomial verifiability.

The core of the proof to SPFP's NP-hardness is this: any subset sum-instance can be *transformed* to a SPFP-instance by creating a corresponding same-value greyscale polygon $p_i$ for every integer $v_i$, and choosing a pixel within a target image such that $t_{pix} = t$. A polygon's restricted greyscale value mutations are multiplications by $2^k$, in which $k$ corresponds to the $k^{th}$ chosen polygon. For our example, this leads to the following transformation

$$\{13, 17, 21, 23\} \rightarrow \begin{pmatrix} 26 \\ 52 \\ 104 \\ 208 \end{pmatrix} \begin{pmatrix} 34 \\ 68 \\ 136 \\ 272 \end{pmatrix} \begin{pmatrix} 42 \\ 84 \\ 168 \\ 336 \end{pmatrix} \begin{pmatrix} 46 \\ 92 \\ 184 \\ 368 \end{pmatrix}$$

4        D. van den Berg

in which the polygons' admissible greyscale values are vertically aligned, a transformation that by all means can be made in (quadratic) polynomial time. Now instead of selecting some values $v_1...v_m$ from set $S$, the task is selecting polygons $w_1...w_m$, and for each $k^{th}$ selected polygon, mutating its color by a factor $2^k$. Note that this transformation allows for the commutativity in subset sum, which is absent in SPFP. After selecting the polygons, the resulting value can be calculated as analogous to Equation 4 for alpha compositing,

$$\frac{1}{2} \cdot w_{1,1} + \frac{1}{4} \cdot w_{2,2} + ... + \frac{1}{2^m} \cdot w_{m,m}. \tag{5}$$

After choosing (the right) polygons and comparing the result from Eq.5 to target pixel $t_{pix}$, any solution can be polynomially transformed back to subset sum, by just taking the original greyscale values of the selected polygons, and dividing them by two. Thereby, any subset sum instance with a set of $n$ integers can be transformed in polynomial time to a single-pixel approximation SPFP-instance with $n$ polygons of $\alpha = 0.5$, and its solution can be transformed back, again in polynomial time. It follows that any algorithm that solves SPFP in time $O(f(x))$ can also solve subset sum in $O(f(x))$. Since subset sum is NP-hard, $f(x)$ cannot be subexponential, and SPFP must also be NP-hard.

## 3   Discussion & Acknowledgments

The original PFP-problem's mutable color and opaqueness, topological constraints, image dimensions and drawing indices are all likely of influence on the problem's hardness. How exactly, still remains to be quantified. Gratitude goes out to Tim Doolan (UvA) for constructive feedback and to Arne Meijs (UvA) for helping out with figure 2.

# Plant Propagation Parameterization: Offspring & Population Size

Marleen de Jonge[0000−0003−4911−2647]
Daan van den Berg[0000−0001−5060−3342]

Informatics Institute, University of Amsterdam
m.r.h.dejonge@student.uva.nl, d.vandenberg@uva.nl

**Abstract.** We investigate a wide range of interdependent parameter values for the number of offspring and the population size in the plant propagation algorithm. An 'optimal window' of parameter values is found, for which the algorithm performs substantially better on five benchmark test functions. Moreover, apart from being within or outside the window, values appear to be largely interchangable, making the algorithm largely independent from specific settings of these parameters.

**Keywords:** Plant Propagation Algorithm · Evolutionary Algorithms · Parametrization · Metaheuristics · Combinatorial Optimization

## 1 Introduction

In recent years, a booming interest has emerged for the implementation of nature-inspired evolutionary algorithms on combinatorial optimization problems. However, the large variety of new algorithms is often not tested thoroughly, and therefore "threatening to lead the area of metaheuristics away from scientific rigor" [7]. As one example, the vast number of possible parameter configurations often form new combinatorial optimization problems in themselves, and determining the 'perfect' settings thereby, becomes a challenging task. Unfortunately, relatively little analysis is done on optimal parameterization in many evolutionary algorithms, despite being "essential for good algorithm performance" [6].

One such algorithm is the Plant Propagation Algorithm (PPA), introduced by Abdellah Salhi and Eric S. Fraga in 2011, which works well on a broad variety of benchmark functions, as well as discrete problems such as the traveling salesman problem, university timetabling and even artistic optimization tasks [4] [8] [5] [1] [3]. After randomly initializing *popSize* initial individuals, the objective values $f(x)$ are calculated, and normalized to [0,1] by $z(x_i) = \frac{f(x_{max})-f(x_i)}{f(x_{max})-f(x_{min})}$ after which the hyperbolic tangent $F(x_i) = \frac{1}{2}(tanh(4 \cdot z(x_i)-2)+1)$ is applied to "[provide a means of emphasising further better solutions over those which are not as good]" [4]. Then, the number of offspring per individual is proportional to its fitness as $n(x_i) = \lceil n_{max} F(x_i) r \rceil$, whereas their mutability is *inversely* proportional as $d_{r,j} = 2(1 - F(x_i))(r - 0.5)$, in which $j$ is the respective dimension; $r$ is a different random number for both equations. Fitter individuals will thereby spawn relatively many offspring with smaller mutations, whereas unfitter individuals produce fewer offspring with larger mutations. In the final step, the offspring get added to the population, from which $popSize$ fittest individuals constitute the next generation.

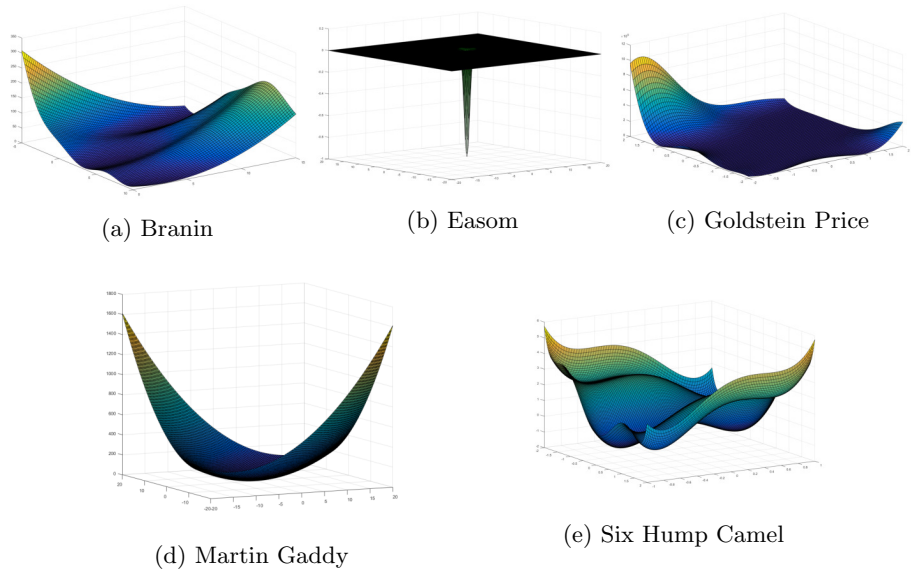

Fig. 1: The five 2D benchmark test functions on which the different parameter values of $popSize$ and $n_{max}$ were tested during a minimization task.

As Salhi and Fraga themselves point out in their seminal work "[parameter values chosen appear suitable for the problems investigated, [but] little analysis has been performed to understand the impact of these parameters]" [4]. Here, we perform a first investigation into two interdependent parameters: the population size ($popSize$) and the maximum number of offspring per individual ($n_{max}$). We explore 400 parameter settings of PPA on five different continuous 2D benchmark test functions (Fig.1) from the original study.

## 2   Experiment & Results

To rigorously test the two parameters, the algorithm was ran with 400 different parameter combinations on all five benchmark test functions with their global minimum set to zero. We choose $1 \leq popSize \leq 40$, $1 \leq n_{max} \leq 10$, and the median best objective value of ten separate runs of the algorithm was taken for each parameter combination (Fig. 2). One run consisted of 10,000 evaluations, amounting in a total of 200 million function evaluations for the whole experiment. It should be noted though, that $popSize$ and $n_{max}$ are interdependent in the number of offspring per generation, so these parameter settings are expected to bring along different numbers of generations per combination.

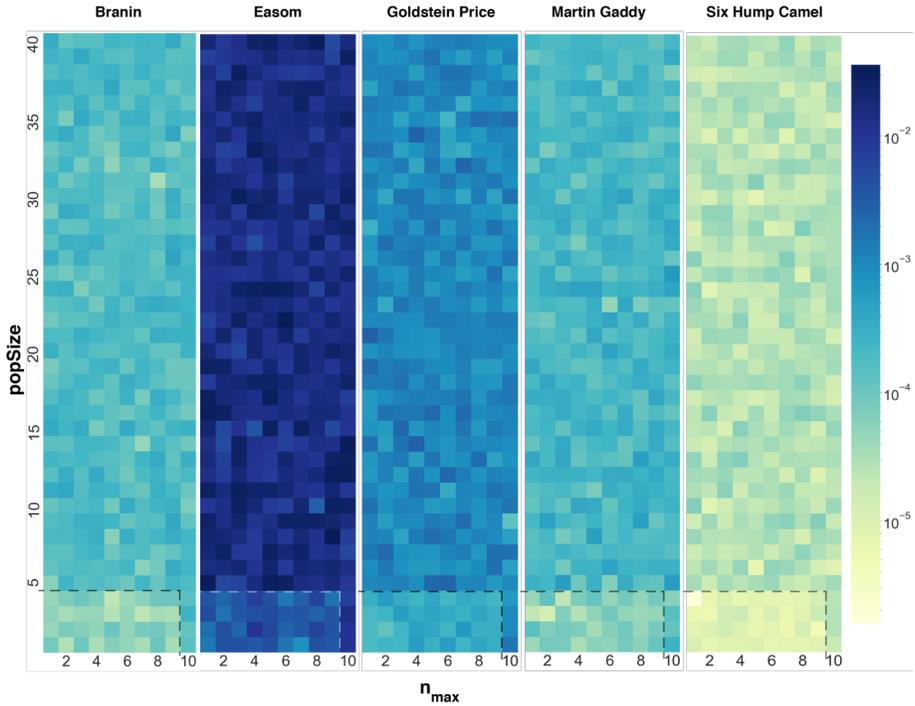

Fig. 2: Results for a wide range of parameter values for *popSize* and $n_{max}$ in PPA on five different benchmark test functions with a zero-normalized global minimum. Each cell represents the median best objective value over 10 runs of 10,000 function evaluations. A consistent window of parameter values (dashed lines) provides significantly better results for all five functions.

The results show a window of parameter configurations that consistently work better than others. For all five of the tested benchmark functions, parameter values $1 \leq popSize \leq 4$ and $1 \leq n_{max} \leq 9$ result in significantly better found objective values (Table 1). But moreover, the differences in standard deviation are small both inside and outside the window, which leaves room for a tantalizing point of discussion.

## 3  Discussion & Future work

Ceteris paribus, it looks like the plant propagation algorithm is largely unsensitive to parameter settings. As can be seen from the heatmap, the only real difference appears whether one chooses a setting inside the window or outside the window, as for both these subsets, their respective standard deviation on the best performance is low.

|  | Inside window | | Outside window | |
|---|---|---|---|---|
| **Function** | $\mu$ | $\sigma$ | $\mu$ | $\sigma$ |
| **Branin** | 5.79e-05 | 2.46e-05 | 1.94e-04 | 7.70e-05 |
| **Easom** | 3.83e-03 | 1.84e-03 | 1.66e-02 | 6.87e-03 |
| **Goldstein Price** | 3.41e-04 | 1.25e-04 | 1.11e-03 | 4.53e-04 |
| **Martin Gaddy** | 6.52e-05 | 2.94e-05 | 2.34e-04 | 9.57e-05 |
| **Six Hump camel** | 8.90e-06 | 3.22e-06 | 2.96e-05 | 1.22e-05 |

Table 1: Overview of the mean and standard deviation per benchmark test function, both within and outside the window of optimal configurations. For each of the functions, the mean best value inside is better than outside this window range. But the standard deviations are low, so apart from the inside-outside-window choice, the algorithm is quite robust against different parameter settings.

Whether these results are consistent across a broader range of benchmark test functions, or more real-worldly problems such as timetabling or the traveling salesman problem, remains to be seen. Finally, we encourage other teams to check, replicate and extend our results, either with or without our publicly accessible repository [2].

# Parallel Microevolutionary Genetic Algorithm for Flexible Job Shop Scheduling Problem

Pedro Coelho[1(✉)][0000-0002-3715-6012], Cristóvão Silva[1][0000-0002-7693-9570]

[1] Univ Coimbra, CEMMPRE, Department of Mechanical Engineering
Pinhal de Marrocos, Coimbra 3030-788, Portugal
`pedro.coelho@dem.uc.pt`

**Abstract.** The Flexible Job Shop Problem is one of the hardest problems in the area of combinatorial optimization. Genetic Algorithms are among the methods of resolution. Despite their natural parallelism, few studies exploit the potential of their parallelization in the emerging powerful Graphics Processing Units. This paper presents a conceptual model for a Parallel Microevolutionary Genetic Algorithm. Taking advantage of the Computer Unified Device Architecture, this conceptual model focuses on thread blocks organization and memory management. Future implementations may validate the potential for good results and its use to tackle more complex extensions of the problem.

**Keywords:** Flexible Job Shop Problem, Parallel Genetic Algorithm, CUDA.

## Introduction

The flexible job shop problem (FJSP), is a generalization of the classic job shop problem (JSP). It is strongly NP-hard and is considered one of the hardest problems in the area of combinatorial optimization. It can be presented as follows. There is a set of jobs, consisting of a number of operations, to be processed in a given order. Each operation must be processed in a machine from a possible set of machines. Each machine can just process one operation at a time, and it cannot be interrupted [1]. The problem consists in assigning operations to machines and sequencing them such that an objective function is optimized.

Several solving methods have been proposed for this problem, ranging from fast and straightforward dispatching rules to sophisticated branch and bound algorithms. In the last years, metaheuristics, in particular, genetic algorithms (GA), have stood out due to their efficiency. GA provides reasonable results, especially when hybridized with other methods [1]. Due to GA natural parallelism, parallel implementation is considered as one of the most promising choices to make them faster [2]. Some studies have explored the parallel processing architectures but without taking full advantage of the massively parallel threads provided by the current Graphics Processing Units (GPU) hardware. An in-depth understanding of GPUs architecture and graphics processing is one of the requirements to adapt/(re)design algorithms to GPUs parallel processing. The introduction of the Computer Unified Device Architecture (CUDA) development environment, by NVIDIA, made it easier for researchers to use GPUs processing power as general-purpose processors.

In this study, a conceptual model to take advantage of the CUDA environment to solve the FJSP is proposed: a parallel microevolutionary GA. This model aims to implement a parallel GA that takes advantage of CUDA threads organization and memory management to solve FJSP.

**Parallel Microevolutionary Genetic Algorithm**

During the process of designing a parallel algorithm for GPU architecture, there is the need to fully exploit the hundreds of thousands of threads running simultaneously [3]. The proposed model exploits that power following two fundamental principles: exposing sufficient parallelism and to enhance memory access.

In the CUDA environment, kernels are the functions that run in the device (GPUs). Only one kernel runs at a time but is executed by multiple threads. Threads are grouped in blocks. All the threads have access to the device high latency Global Memory, but threads in the same block can access a low latency shared memory. However, the block shared memory is limited to 48 KB while the global memory is in the order of GB. This organization provides the basis for this conceptual model.

According to Goldberg [4], a GA with a small population of three individuals is sufficient to converge. This proved convergence makes it viable to propose a conceptual model based on a solo island approach where a small population undergoes microevolution. After a certain period of microevolution, individuals are mixed in a pool and redistributed through islands. A reasonable solution is expected after several microevolution cycles. Figure 1 outlines the conceptual model. In the CUDA environment, this solo island is seen as a thread block, accessing mainly the low latency shared memory, where the population chromosomes data is stored. During the algorithm implementation, data parallelism is exposed by assigning one thread to process one gene of the chromosome. Therefore, gene level granularity is obtained.

The algorithm model starts by copying data from the Central Processing Unit (CPU) memory to the GPU Global Memory and launches a single thread kernel. This main kernel takes advantage of the CUDA Dynamic Parallelism to launch a multi-block thread kernel to generate the initial population. Each thread block generates one individual chromosome, and each thread is responsible for generating a chromosome gene. Genes are stored directly in the global memory.

Chromosomes data is organized as a structure of arrays, allowing faster memory access by threads. The problem solution is represented by a double operation-based chromosome, where each index corresponds to a job operation. One part of the chromosome encodes machine assignment, and the other encodes operations sequence. The assignment genes are generated, with their thread choosing, randomly, one possible machine to each operation. The sequence genes are generated, with their thread assigning randomly one number between 0 and 1 to the corresponding operation. The use of random keys representation requires a sort-by-key algorithm in the decoding process, but it allows the use of the same crossover operator for the entire chromosome. The chromosome has an extra gene to store the results of performance evaluation that is initialized randomly.

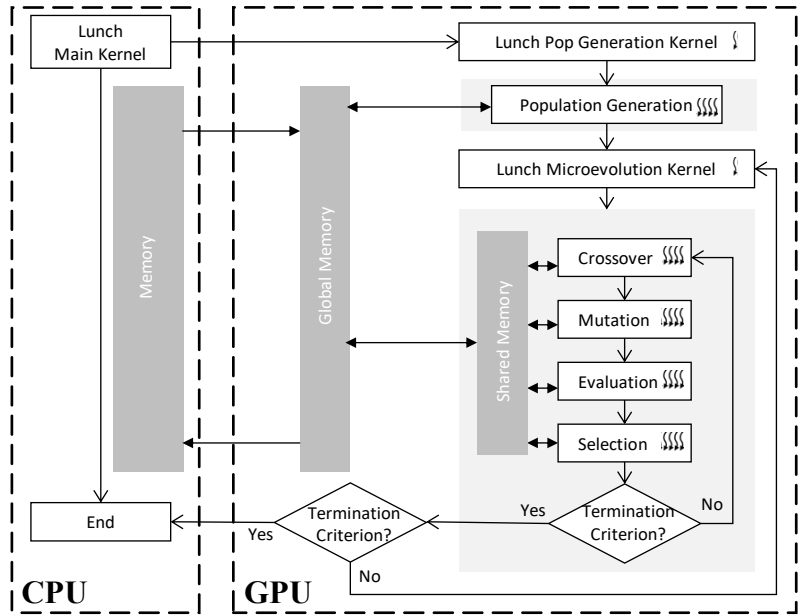

**Fig. 1.** The Parallel Microevolutionary Genetic Algorithm Conceptual Model.

After the initial population generation, the main kernel restarts running and launches the microevolution kernel. This kernel runs in several thread blocks. The first step consists to copy the jobs operation data and three chromosomes to the shared memory. The next steps implement the GA. The microevolution occurs while the kernel cycles through the GA operators a pre-defined N number of times.

The microevolution starts with the crossover operator. An n-point crossover mixes the best-evaluated individual with the two others (resulting in 7 individuals). The number of crossover points is a function of the number of jobs operations. Each thread copies one gene from the parents to the offspring chromosome storing them it in the shared memory. The n-point crossover enhances memory coalescing.

In the mutation operator, each thread generates a random number and mutates its gene if the obtained number is higher than a pre-defined threshold point. In the machine assignment genes, threads change the gene value by another randomly chosen from the possible ones. In the machine sequence genes, threads change the gene value by generating another random value between 0 and 1.

Due to FJSP complexity, evaluation operator is the most time-consuming function. The development of an efficient algorithm for this operator is crucial. The chromosome decoding starts by sort operations based on their random key and assigns them a sequence order. Combining machine assignment information with the operation sequence, the sequence of the operations on each machine is provided. At this point, the FJSP becomes a JSP. The next step is to construct the scheduling, determining the starting time of each operation. The schedule allows computing the value of the objective function to be optimized. The individual performance is registered in the corresponding gene. A first approach to this operator may be the parallel implementation presented by Bozejko *et al.* [5]. This author models the JSP problem as a disjunctive graph and adopts the Floyd-Warshall algorithm for the longest path between each pair of nodes.

The selection operator ranks the chromosomes performance values, and selects the three individuals with better performance as parents for the next generation. However, in the last cycle, the selected chromosomes are copied back to global memory and microevolution kernel ends.

The main kernel restarts running, relaunching the microevolution kernel until the termination criterion is reached. At that point, the population is ranked by performance, the main kernel ends and the CPU retrieves the best solution from the global memory.

## Conclusions

In this paper, a parallel microevolutionary GA conceptual model is presented. Aiming to solve the FSJP, the model adjusts the CUDA environment to the problem specificities. To achieve proper results, it is crucial to fine-tune the GA parameters. The total jobs operation is also an essential parameter to achieve efficient CUDA memory management. Future implementations may validate its potential for good results and its use to tackle more complex extensions of the problem.

**Acknowledgements**

This research is sponsored by FEDER funds through the program COMPETE – Programa Operacional Factores de Competitividade – and by national funds through FCT – Fundação para a Ciência e a Tecnologia –, under the project UID/EMS/00285/2020 and the doctoral grant to P.C. (SFRH/BD/129714/2017).